\documentclass[runningheads,a4paper]{llncs}

\usepackage{mathtools}
\usepackage{multirow}
\usepackage{adjustbox}

\usepackage{graphicx}
\usepackage{subfig}
\usepackage{amssymb}

\usepackage{url}
\urldef{\mailsa}\path|{kangmin89, dinara_aliyeva, bj1123, yalphy}@korea.ac.kr|
\newcommand{\keywords}[1]{\par\addvspace\baselineskip
\noindent\keywordname\enspace\ignorespaces#1}

\begin{document}

\mainmatter  

\title{Incorporating Word Embeddings into Open Directory Project based Large-scale Classification}

\titlerunning{Incorporating Word Embeddings into ODP-based Large-scale Classification}

%
%
\author{Kang-Min Kim%
, Aliyeva Dinara, Byung-Ju Choi, SangKeun Lee
}
\authorrunning{}

\institute{Korea University, Seoul, Republic of Korea\\
\mailsa\\
\url{}}

%
%

\toctitle{Incorporating Word Embeddings into Open Directory Project based Large-scale Classification}
\tocauthor{TBD}
\maketitle

\vspace*{-0.1cm} 

\begin{abstract}
Recently, implicit representation models, such as embedding or deep learning, have been successfully adopted to text classification task due to their outstanding performance. However, these approaches are limited to small- or moderate-scale text classification.
Explicit representation models are often used in a large-scale text classification, like the Open Directory Project (ODP)-based text classification. However, the performance of these models is limited to the associated knowledge bases. In this paper, we incorporate word embeddings into the ODP-based large-scale classification. To this end, we first generate category vectors, which represent the semantics of ODP categories by jointly modeling word embeddings and the ODP-based text classification. We then propose a novel semantic similarity measure, which utilizes the category and word vectors obtained from the joint model and word embeddings, respectively. The evaluation results clearly show the efficacy of our methodology in large-scale text classification. The proposed scheme exhibits significant improvements of 10\% and 28\% in terms of macro-averaging F1-score and precision at $k$, respectively, over state-of-the-art techniques.
\end{abstract}

\section{Introduction}


Text classification is the process of determining and assigning topical categories to text. It plays an important role in many web applications, such as contextual advertising  \cite{Lee:MC-TWEB}, topical web search \cite{Broder:Robust}, and web search personalization \cite{Chirita:Meta}. Usually, text classification requires a sufficiently large taxonomy of topical categories to capture various topics in arbitrary texts. In addition, it is necessary to collect a large amount of training data for each category in the taxonomy.

Many studies have utilized an implicit representation model \cite{Wang:Short}, such as embedding \cite{Mikolov:Word,Mikolov:Phrase,Le:Sentence} or a deep neural network \cite{Kim:CNN}, which adopts dense semantic encodings and measures semantic similarity accordingly. Implicit representation models have been successfully adopted for text classification task. Such implicit representation models, however, may perform poorly in a large-scale text classification (as we shall show in Section 5.4). This is largely attributed to the fact that the training data for each category is relatively insufficient and distributed unevenly among classification categories. In addition, such approaches are not intuitively interpretable to humans.

In another line of work, many studies have been done with an explicit representation model \cite{Wang:Short}, which uses popular knowledge bases, such as ProBase, Wikipedia, or the Open Directory Project (ODP)\footnote{http://www.curlie.org}. Because the explicit model represents knowledge in terms of vectors that are interpretable to both humans and machines, it is relatively easy for humans to tune and understand it. Another advantage of the explicit representation model is that it enables a large-scale text classification with the direct representation of a large-scale knowledge taxonomy already built-in.  

To handle the large-scale text classification, several works \cite{Lee:MC-TWEB,Shin:Wikipedia,Ha:MCAD} have utilized the ODP, which is a large-scale and taxonomy-structured web directory. These studies have used their explicit representation of text to represent ODP knowledge, based on bag-of-words \cite{Lee:MC-TWEB,Ha:MCAD} or bag-of-phrases \cite{Shin:Wikipedia} to develop ODP-based text classification techniques. They showed that ODP-based text classification techniques are effective at the large-scale text classification. The performance of previous ODP-based text classification, however, is limited to ODP and/or Wikipedia knowledge bases. 

To alleviate the limitation of ODP-based text classification, we incorporate word embeddings into the ODP-based text classification. To this end, we propose two novel joint models of ODP-based classification and word2vec, a representative word embeddings technique. The joint models seek to project both words and ODP categories into the same vector space. Therefore, category vectors of ODP categories successfully identify words learned from external knowledge. In addition, we effectively measure the semantic relatedness between an ODP category and a document by utilizing both category and word vectors. In summary, our contributions are three-fold:
\begin{itemize} 
\item We propose a novel methodology to handle the large-scale text classification, which utilizes both the explicit and implicit representation. 
\item We develop two novel joint models of ODP-based classification and word2vec to generate category vectors that represent the semantics of ODP categories. In addition, we develop a new semantic similarity measure that utilizes both the category and word vectors.
\item We demonstrate the efficacy of the proposed methodology through extensive experiments on real-world datasets. The performance evaluation clearly shows that our approach significantly outperforms the state-of-the-art techniques in terms of macro-averaging F1-score and precision at $k$.
\end{itemize}

The remainder of this paper is organized as follows. We briefly describe the ODP-based knowledge representation and word2vec in Section 2. Section 3 describes the joint models of ODP-based classification and word2vec to generate category vectors. Section 4 details the similarity measure between a category and document. We present the performance evaluation results in Section 5. We discuss related research and conclude this work in Sections 6 and 7, respectively.

\section{Preliminary}
\subsection{ODP-based Knowledge Representation}
We employ the ODP-based text classification scheme \cite{Lee:MC-TWEB} as our explicit representation model.
To compute the centroid $\overrightarrow{\mu_i}$ of category $c_i$, we calculate the averaged term vector of all ODP documents as:
\begin{equation}
\label{centroid}
\begin{split}
\overrightarrow{\mu_i} & =\frac{1}{\|D_{c_i}\|} \sum_{d \in D_{c_i}}\overrightarrow{d}
\end{split}
\end{equation}

 where $D_{c_i}$ is a set of ODP documents in $c_i$, and $\overrightarrow{d}$ is a weighted vector represented as a {\it tf-idf} value. Due to the large-scale taxonomy structure of the ODP, however, each ODP category contains a different number of documents, sometimes resulting in sparsity or unavailability of training documents in a category. This issue is addressed in the works \cite{Lee:MC-TWEB,Ha:MCAD}, in which they merge the centroid $\overrightarrow{\mu_i}$ of the descendant categories to build a classifier. As a result, this approach outperforms all other ODP-based text classifiers, and exhibits a stable performance in large-scale text classification \cite{Lee:MC-TWEB,Ha:MCAD}. Therefore, we utilize this approach to compute the centroid $\overrightarrow{\mu_i}$ of category $c_i$. 

\vspace*{-0.6cm} 
   
\begin{table}[]
\centering
\caption{Example of ODP-based Representation. A document $d$, ``{\it Trump became prez}", to be classified, and a category $c_1$, {\it Society$/$}{\it Government$/$}{\it President} are considered.}
\label{limitexample}
\renewcommand{\arraystretch}{0.9}
\begin{tabular}{|c|ccccc|}
\cline{2-6}
\multicolumn{1}{l|}{} & \multicolumn{5}{c|}{\textbf{term weights}}        \\ \hline
\textbf{vector}	& trump	& president	& prez	& government	& ...	\\ \hline\hline
term vector of $d$	& 0.67	& 0	& 0.51	& 0 & ...	\\ \hline
centroid vector of $c_1$	& 0.10	& 0.44	& 0.05	& 0.31	& ...\\ \hline
\end{tabular}
\end{table} 
 
\vspace*{-0.4cm} 
   
For example, as shown in Table \ref{limitexample}, the category $c_1$, {\it Society$/$}{\it Government$/$} {\it President} is explicitly represented by the centroid vector. Given a document $d$, however, ``{\it Trump became prez}'', the ODP-based classification may not be able to classify the document $d$ as the category $c_1$. This is because, this approach cannot capture the semantic relations between words (e.g., {\it prez} and {\it president}).

\subsection{Word2Vec}
To complement the ODP-based classification, we adopt the word2vec \cite{Mikolov:Word,Mikolov:Phrase}, a popular word embeddings technique. In word2vec, each word vector is trained using a shallow neural networks language model, such as continuous bag-of-words (CBOW) or Skip-gram \cite{Mikolov:Word}. Skip-gram aims to predict context words given a target word in a sliding window. Mathematically, given a sequence of training words $w_1, w_2, w_3 ..., w_T$,  the objective of Skip-gram is to maximize the following average log probability:
\begin{equation}
\label{skip-gram}
\begin{split}
\frac{1}{T} \sum^{T}_{i=1}\sum^{i+k}_{c=i-k,{c}\neq{t}} log \: {p(w_c|w_t)}
\end{split}
\end{equation}

where $k$ is the size of the context window centered at the target word, and $w_t$ and $w_c$ are the target and context words, respectively.

Trained word vectors with similar semantic meanings would be located at high proximity within the vector space. For example, the word vectors of {\it president} and {\it prez} would be located close to each other. On the other hand, the word vectors of {\it president} and {\it casino} would be located much more distantly in the embedding space. In addition, word vectors can be composed by an element-wise addition of their vector representations, e.g., {\it Russian} + {\it river} = {\it Volga River}. This property of the vectors is called ``additive compositionality" \cite{Mikolov:Phrase}. Due to the simple structure of word2vec, many previous studies have proposed variants of the word2vec model to go beyond the word-level to achieve document-, topic-, or concept-level representations \cite{Le:Sentence,Cheng:WCS}.

\section{Joint Models of Explicit and Implicit Representation}
In this section, we describe two joint models of ODP-based text classification and word2vec. These joint models generate category vectors, which represent the semantics of ODP categories. Each category vector not only semantically encodes the explicitly expressed ODP category, but also understands semantically related words that do not appear in the ODP knowledge base. This is because they are projected into the same semantic space as word vectors learned in an additional volume of knowledge outside the ODP.

\subsection{Generating Category Vector with Algebraic Operation}
Given the centroid vector of an ODP category and word vectors of the pre-trained word2vec model, our first approach generates the category vector by using the vector scalar multiplication and vector addition methods, as follows. 

First, we multiply the term weights of each word in the ODP category by each word vector of the words. Second, the weighted word vectors are composed as a category vector using element-wise addition. This type of vector algebra is quite simple, but it can also clearly represent the semantics of an ODP category. This is because word vectors are not only multiplied by a precisely trained term weight from the centroid vector, but also have additive compositionality. 

\begin{figure}[t]
\includegraphics[width=4.8in]{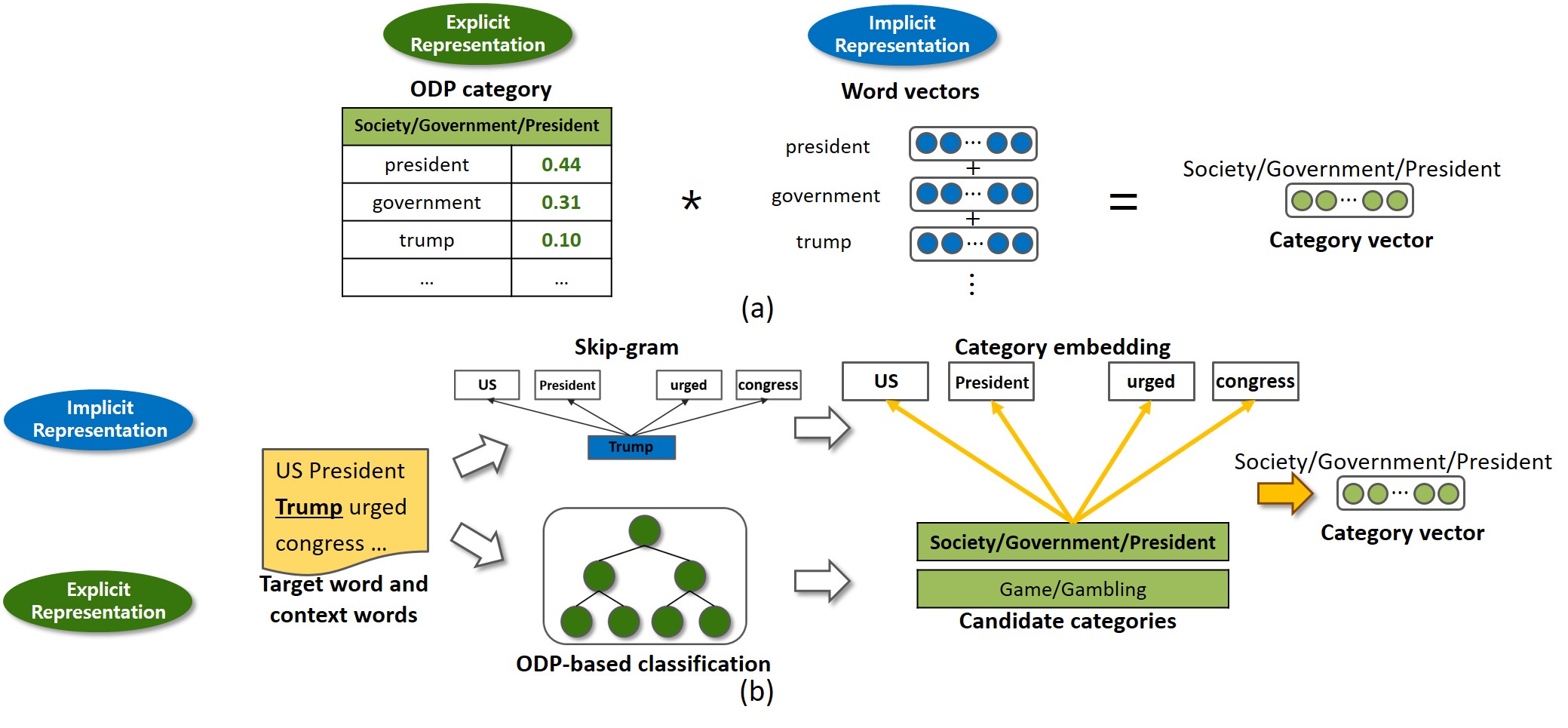}
 \centering
 \caption{Illustration of Category Vector Generation with Algebraic Operation (a) and Embedding (b)}
 \label{fig:categoryVector}
 \vspace*{-0.2cm} 
\end{figure}

The logic for generating the category vector of the ODP category is as follows:
\begin{equation}
\label{topic vector(category)}
\begin{split}
\overrightarrow{C_i} & =\sum_{w \in W_i} \overrightarrow{\mu_i}(w) \cdot \overrightarrow{w}
\end{split}
\end{equation}

 where $\overrightarrow{C_i}$ is the category vector of $c_i$, $W_i$ is the set of words of $c_i$, $\overrightarrow{w}$ is the word vector (obtained from the pre-trained word2vec model) of word $w$, and $\overrightarrow{\mu_i}(w)$ is the term weight of $w$ in $c_i$. For example, in Figure \ref{fig:categoryVector}(a), the word vectors of $president$, $government$, and $trump$ are multiplied by 0.44, 0.31, and 0.10, respectively, then the weighted word vectors are added. Finally, we obtain the category vector of the category {\it Society$/$}{\it Government$/$}{\it President}. Vector representations of documents to be classified are generated in the same manner.

\subsection{Generating Category Vector with Embedding}
Our second approach extends word2vec to represent category vectors, instead of using the pre-trained word2vec model to compose word vectors in ODP categories. We first assign appropriate ODP categories for each word in a text corpus. Then, we train the category vectors of the assigned ODP categories by applying a modified Skip-gram model. The category vector of an ODP category is expected to represent the collective semantics of words under this category.

The process of generating category vectors with embedding is as follows. First, we identify candidate ODP categories for the target word. If an ODP category is largely associated with the target word, the ODP-based text classification selects this category as a candidate. The ODP-based text classification determines the degree of association by using the term weight of the target word in each ODP category. For example, when $Trump$ is the target word, the ODP-based classification identifies categories such as {\it Game$/$Gambling} and {\it Society$/$Government$/$}{\it President}, as shown in Figure \ref{fig:categoryVector}(b). We then select the most appropriate ODP category in the current context by using the ODP-based text classification. For example, when the context is {\it``US President Trump urged congress"}, the most appropriate category is {\it Society$/$}{\it Government$/$}{\it President}. Finally, we apply the modified Skip-gram algorithm, which trains the category vector corresponding to the most appropriate category.

The objective of category embedding is to maximize the following average log probability:
\vspace*{-0.3cm}
\begin{equation}
\label{topic_embedding}
\begin{split}
\frac{1}{T} \sum^{T}_{i=1}\sum^{i+k}_{c=i-k,{c}\neq{t}} log \: {p(w_c|w_t)p(w_c|c_t)}
\end{split}
\end{equation}

Unlike the Skip-gram model, where the target word $w_t$ is used only to predict context words, the category embedding model also uses the ODP category $c_t$ of the target word to predict context words. 

\section{Semantic Similarity Measure}
We develop a novel semantic similarity measure, on the basis of category and word vectors, which captures both the semantic relations between words and the semantics of ODP categories. 

\subsection{Using Word-level Semantics}
First, we propose a semantic similarity measure that considers word-level semantics by using only the word vectors. The word vectors can be used to calculate the semantic relatedness between two words. The key idea of this measure is to align words with similar meanings in a category and document, although the words represented in this category and document are different. 

Before describing the proposed measure, we explain how to compute the similarity between category $c_i$ and document $d$ by means of the existing ODP-based text classification as follows:
\begin{equation} \label{cosine_sim}
\text{cos}(c_i \text{, }d)= \frac{\sum^{n_{c_i}}_{j=1}\sum^{n_d}_{k=1}\delta(w_j - w_k) \cdot  \overrightarrow{\mu_i}(w_j) \cdot \overrightarrow{d}(w_k)}{\|\overrightarrow{\mu_i}\|\cdot\|\overrightarrow{d}\|}
\end{equation}

where $w_j$ and $w_k$ are non-zero terms in centroid vector $\overrightarrow{\mu_i}$ of $c_i$ and term vector $\overrightarrow{d}$, respectively, while $n_{c_i}$ and $n_d$ are the number of non-zero terms in $\overrightarrow{\mu_i}$ and $\overrightarrow{d}$, respectively. $\delta(\cdot)$ is the Dirac function defined by $\delta(0)$ = 1 and $\delta(other)$ = 0 \cite{Song:Dense-ESA}.

The cosine similarity between the centroid vector of category and the term vector of document could increase when $w_j$ and $w_k$ are equal. However, in Table \ref{limitexample}, we observe that {\it prez} has a very similar meaning to {\it president}, which is a very important word in the category {\it Society$/$Government$/$}{\it President}. Therefore, we propose a new measure that increases the similarity between proper $\overrightarrow{\mu_i}$ and $\overrightarrow{d}$ by utilizing word2vec. By substituting the Dirac function $\delta(\cdot)$ with the word similarity $\phi(\cdot)$, it is possible to consider semantic relatedness between two words and calculate the weight more densely:
\begin{equation} \label{augmentation}
\text{sim}(c_i \text{, }d) = \frac{\sum^{n_{c_i}}_{j=1}\sum^{n_d}_{k=1}\phi(w_j, w_k) \cdot \overrightarrow{\mu_i}(w_j) \cdot \overrightarrow{d}(w_k)}{\|\overrightarrow{\mu_i}\|\cdot\|\overrightarrow{d}\|}
\end{equation}

where $\phi(\cdot)$ is the word similarity function. Given two words $w_j$ and $w_k$, we define the word similarity function $\phi(w_j, w_k)$ in Eq. (\ref{augmentation}) as follows:
\begin{equation}
\label{word similarity}
\begin{split}
\phi(w_j, w_k) & =\begin{cases}
               cos(\overrightarrow{w_j}, \overrightarrow{w_k}) \text{ if } cos(\overrightarrow{w_j}, \overrightarrow{w_k})> \theta,\\
              \text{0} \qquad\qquad \text{  otherwise}\\
            \end{cases}
\end{split}
\end{equation}

where $\overrightarrow{w_j}$ and $\overrightarrow{w_k}$ are the word vectors of $w_j$ and $w_k$, $cos(\overrightarrow{w_j}, \overrightarrow{w_k})$ is the cosine similarity between $\overrightarrow{w_j}$ and $\overrightarrow{w_k}$, and $\theta$ is a threshold, which is empirically set to 0.6 in our analysis. The similarity between $\overrightarrow{\mu_i}$ and $\overrightarrow{d}$ increases not only when $w_j$ and $w_k$ are equal, but also they have similar semantics. For example, {\it prez} and {\it president} have highly similar semantics in Table \ref{limitexample}. The semantic similarity using word-level semantics, thus, is additionally computed by 0.51 $\times$ 0.44 $\times$ $\phi(prez, president)$, unlike the original cosine similarity.

\subsection{Using Category- and Word-level Semantics}
In this paper, we develop a robust similarity measure by utilizing both the category and word vectors. A category vector is utilized as a pseudo word in the process of computing semantic similarity. A new measure can be expressed as follows:
\begin{equation} \label{augmentation2}
\text{sim}'(c_i \text{, }d) = \frac{\sum^{n_{c_i+1}}_{j=1}\sum^{n_d+1}_{k=1}\phi(w_j, w_k) \cdot \overrightarrow{\mu_i}(w_j) \cdot \overrightarrow{d}(w_k)}{\|\overrightarrow{\mu_i}\|\cdot\|\overrightarrow{d}\|}
\end{equation}

In Eq. (\ref{augmentation2}), the category vector is inserted into the corresponding category as the $(n_{c_i}+1)^{th}$ word. This is motivated by the fact that category vectors share the same semantic space with word vectors. Similarly, the document vector is inserted into the corresponding document as the $(n_d+1)^{th}$ word. We will examine how to insert a category vector as a pseudo word by determining the weight (i.e., pseudo term weight) $\alpha$ of the category vector through many parameter experiments in Section 5.4.

\section{Experiments}
\subsection{Datasets}
\vspace*{-1.0cm}
\begin{table}[]
\centering
\small
\caption{Statistics of Datasets}
\label{datasets}
\renewcommand{\arraystretch}{0.9}
\begin{tabular}{c|l|c|c}
\hline
\multicolumn{2}{c|}{\textit{}} 			& Training dataset		 & Test dataset\\ \hline \hline
ODP	& No. Categories	& 2,735/13 & 2,735/13\\ \cline{2-4}
(large-scale/moderate-scale)	& No. Webpages	& 52,046/51,856 & 24,121/24,046\\ \hline
\multirow{1}{*}{NYT}    & No. Articles      				& -& 120 \\ \cline{2-3}
 \hline
\end{tabular}
\end{table}

\vspace*{-0.8cm}

\subsubsection{Training Datasets}
We use the RDF dump from the original ODP dataset released on January 8, 2017, which contains 802,379 categories and 3,624,444 webpages. To obtain a well-organized ODP taxonomy, we apply heuristic rules suggested in \cite{Lee:MC-TWEB} and build our own taxonomy with 2,735 categories. Thus, the final training dataset used in our experiments consists of 52,046 webpages. To construct the moderate-scale classification dataset, we use only 13 top-level categories from the ODP taxonomy by excluding two categories, {\it Top$/$News} and {\it Top$/$Adult}, which contain fewer than 100 webpages. Thus, the training dataset used in the moderate-scale classification consists of 51,856 webpages.

In addition to the ODP dataset, we train our category embedding model and word2vec model on the ``One Billion Word Language Modeling Benchmark" dataset released by Google\footnote{https://code.google.com/archive/p/word2vec/}. The word and category vectors are 300-dimensional, while the window size is set to 5 with 15 negative samples.

\subsubsection{Test Datasets}
We build two test datasets, ODP and NYT, to evaluate our methodology. The ODP test dataset consists of webpages collected from the original ODP. The webpages in each category are randomly divided into a training set and a test set at a ratio of seven to three. In particular, we build two kinds of ODP test datasets. In the large-scale classification task, we collect 24,121 webpages from 2,735 ODP categories in our taxonomy, while collecting 24,046 webpages from 13 ODP categories in the moderate-scale classification task. In addition to the ODP test datasets, we select six categories related to the New York Times: {\it art}, {\it business}, {\it food}, {\it health}, {\it politics}, and {\it sports}, as the source for our second test dataset. We randomly collect 20 news articles from each of these categories. Table \ref{datasets} shows the statistics of datasets.
\vspace*{-0.3cm}
\subsection{Evaluation Metrics}
For the ODP test dataset, we use the macro-averaging precision, recall, and $F_1$-score \cite{Yang:Evaluation} as the classification performance metric. We adopt the macro-averaging, which assigns equal weights to each category instead of each test document, because the distribution of the ODP training dataset is highly skewed \cite{Lee:MC-TWEB,Ha:MCAD}. For the NYT test dataset, we use precision at $k$. Three participants manually assess the top-k ODP categories obtained by text classifiers in three scales: relevant, somewhat relevant, and not relevant. 
\vspace*{-0.3cm}
\subsection{Experimental Setup}
We evaluate the performance of six methods. We adopt the ODP-based text classifier for our experiments, because the ODP-based text classification \cite{Lee:MC-TWEB,Ha:MCAD} outperforms many well-known classification methods such as Naive Bayes, k-Nearest Neighbor and Support Vector Machine. Other baselines include the paragraph vector \cite{Le:Sentence} and convolutional neural networks-based text classifier \cite{Kim:CNN}, which are state-of-the-art methods on multi-class text classification. In our experiments, we compare the following methods:
\vspace*{-0.1cm}
\begin{itemize}
\item \text{$ODP$ (baseline)}: This is the ODP-based text classification only \cite{Lee:MC-TWEB}.
\item \text{$PV$ (baseline)}: This is the text classification method using paragraph vectors \cite{Le:Sentence}. The learned vector representations have 1000 dimensions. We represent ODP categories by averaging the document embeddings for each document in a category. We use the cosine similarity to calculate the similarity between a category and document.
\item \text{$CNN$ (baseline)}: This is the convolutional neural networks-based text classifier \cite{Kim:CNN}. The dimension of word embedding is 300, and the number of filters for the CNN is 900. Weights other than the word embedding layer are initialized by the Gaussian distribution, with a mean of 0 and a standard deviation of 0.01. We use the ReLU for nonlinearity. Optimization is performed using SGD with a mini-batch size of 64 with RMSProp for acceleration.  
\item \textbf{$ODP_{CV}$}: This is our proposed text classification method using category vectors, which are generated by the joint model of ODP-based text classification and word2vec. We use the cosine similarity to calculate the similarity between a category and document vector.
\item \textbf{$ODP_{WV}$}: This is our proposed ODP-based text classification combined with the similarity measure of word-level semantics.
\item \textbf{$ODP_{CV+WV}$}: This is our proposed ODP-based text classification combined with the similarity measure of both category- and word-level semantics.
\end{itemize}

\vspace*{-0.3cm}

\subsection{Experimental Results}

We first compare the two methods to generate category vectors with the ODP dataset (2,735 categories). In Table \ref{Comparison_TVs},  $ODP_{CV(Algebra)}$ denotes the text classification utilizing the category vector generated by algebraic operations, while $ODP_{CV(Embedding)}$ denotes the text classification utilizing the category vector generated by embedding. Unexpectedly, we observe that a simple $ODP_{CV(Algebra)}$ clearly outperforms a relatively elaborate $ODP_{CV(Embedding)}$. Thus, we adopt $ODP_{CV(Algebra)}$ in the remaining experiments, which is simply denoted by $ODP_{CV}$.

\vspace*{-0.8cm}

\begin{table}[h]
\centering
\small
\caption{Comparison of Category Vector Generations on the ODP Dataset (2,735 categories)}
\label{Comparison_TVs}
\renewcommand{\arraystretch}{1.0}
\begin{tabular}{c|c|c|c}
\hline
\multicolumn{1}{c|}{\textit{}} 			& Precision		& Recall		 & F1-score\\ \hline \hline
\multicolumn{1}{c|}{$ODP_{CV(Algebra)}$} 	        	& \textbf{0.449}   & \textbf{0.458} & \textbf{0.453}\\ \hline
\multicolumn{1}{c|}{$ODP_{CV(Embedding)}$}	& 0.278   & 0.195	& 0.230	\\ \hline
\end{tabular}
\vspace*{-0.3cm}
\end{table}

\begin{figure}[h]
\includegraphics[width=4.3in]{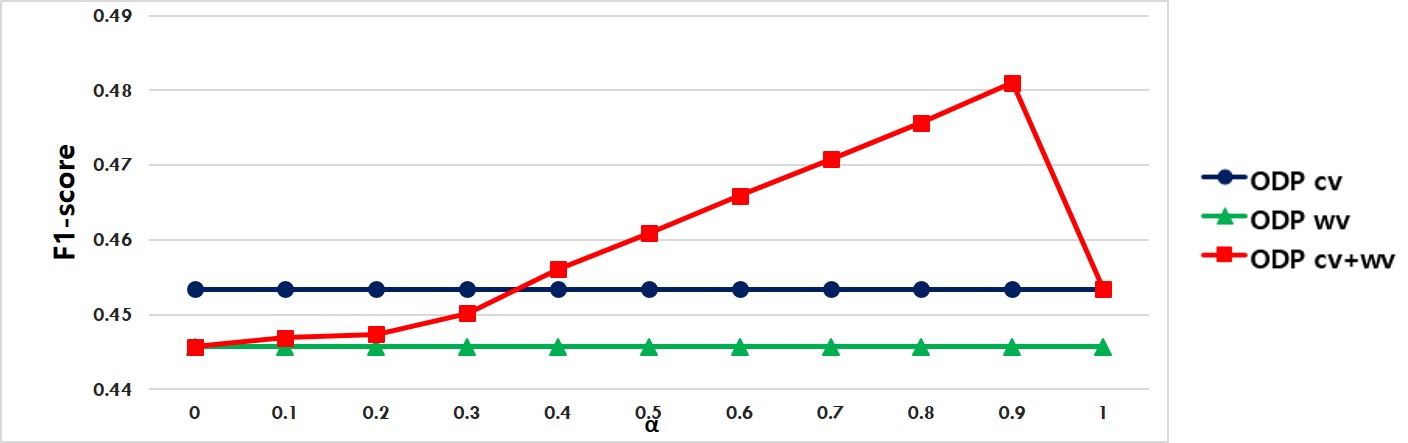}
\centering
\caption{Classification Performance based on Different $\alpha$ Values}
\label{fig:Parameter}
\vspace*{-0.5cm}
\end{figure}

Next, we perform a parameter setting to determine the term weight $\alpha$ of a category vector as a pseudo word. Figure \ref{fig:Parameter} shows the classification performance obtained by $ODP_{CV+WV}$ based on different $\alpha$ values. We find that the curve reaches a peak at $\alpha$ = 0.9. This result shows that the category vector plays a major role in the performance of $ODP_{CV+WV}$. However, we observe that when the weight of category vector is 1.0, the performance drops sharply. This means that the word overlap feature is still helpful. In the remaining experiments, $\alpha$ is set to 0.9 for $ODP_{CV+WV}$.

\vspace*{-0.4cm}

\begin{table}[h]
\centering
\small
\caption{Classification Performance on the ODP Dataset.}
\label{ODP_Performance}
\renewcommand{\arraystretch}{0.90}
\subfloat[large-scale (2,735 categories)]{
\begin{tabular}{c|c|c|c}
\hline
\multicolumn{1}{c|}{\textit{}} 			& Precision		& Recall		 & F1-score\\ \hline \hline
\multicolumn{1}{c|}{$ODP$ \cite{Lee:MC-TWEB}} 	        	& 0.431   & 0.440 & 0.436\\ \hline
\multicolumn{1}{c|}{$PV$ \cite{Le:Sentence}}	& 0.331   & 0.398	& 0.361	\\ \hline
\multicolumn{1}{c|}{$CNN$ \cite{Kim:CNN}}	& 0.402   & 0.232	& 0.294	\\ \hline \hline
\multicolumn{1}{c|}{$ODP_{CV}$}          & 0.449   & 0.458	& 0.453	 \\ \hline
\multicolumn{1}{c|}{$ODP_{WV}$}     		& 0.451   & 0.440	& 0.446	\\ \hline
\multicolumn{1}{c|}{$ODP_{CV+WV}$}       & \textbf{0.468}   & \textbf{0.494}   & \textbf{0.481} \\ \hline
\end{tabular}}
\hspace{.5cm}
\subfloat[moderate-scale (13 categories)]{
\begin{tabular}{c|c|c|c}
\hline
\multicolumn{1}{c|}{\textit{}} 			& Precision		& Recall		 & F1-score\\ \hline \hline
\multicolumn{1}{c|}{$ODP$ \cite{Lee:MC-TWEB}} & 0.667   & \textbf{0.707} & 0.687\\ \hline
\multicolumn{1}{c|}{$CNN$ \cite{Kim:CNN}}	& \textbf{0.736}   & 0.661 & \textbf{0.696}	\\ \hline
\end{tabular}}
\vspace*{-0.7cm}
\end{table}

Table \ref{ODP_Performance}(a) summarizes the experimental results for text classification on the ODP test dataset with 2,735 target classes. We observe that $ODP_{CV+WV}$ outperforms all the other proposed methods, as well as the baselines. $ODP_{CV+WV}$ performs better than $ODP$ over 9\%, 12\%, and 10\% on average in terms of precision, recall, and F1-score, respectively. Our experimental results show that $PV$ \cite{Le:Sentence} performs worse than $ODP$. In addition, it turns out that $CNN$ \cite{Kim:CNN} performs the worst among the six methods. This can be explained by the fact the distribution of webpages is skewed toward a few categories in the original ODP \cite{Lee:MC-TWEB}. Actually, we observe that 73\% of ODP categories contain fewer than five webpages. 

We also compare the performance of $CNN$ with the $ODP$ baseline on the ODP test dataset with 13 target categories. From Table \ref{ODP_Performance}(b), we observe that $CNN$ exhibits a better performance than $ODP$ in the moderate-scale text classification. From Table \ref{ODP_Performance}, we confirm that $CNN$ is indeed limited to the moderate-scale text classification.

\vspace*{-0.6cm}

\begin{table}[h]
\centering
\small
\caption{Classification Performance on the NYT Dataset (2,735 categories).}
\label{NYT_Performance}
\renewcommand{\arraystretch}{0.9}
\begin{tabular}{c|c|c|c|c|c}
\hline
\multicolumn{6}{c}{Precision at \textit{k}}  \\ \hline
\multicolumn{1}{c|}{\textit{k}} 			& 1         & 2		& 3		 & 4          & 5        \\ \hline \hline
\multicolumn{1}{c|}{$ODP$ \cite{Lee:MC-TWEB}} 					& 0.575     & 0.496    & 0.450	 & 0.421      & 0.403    \\ \hline
\multicolumn{1}{c|}{$PV$ \cite{Le:Sentence}}   					& 0.317     & 0.292    & 0.261	 & 0.250      & 0.245    \\ \hline
\multicolumn{1}{c|}{$CNN$ \cite{Kim:CNN}}   					& 0.242     & 0.200    & 0.186	 & 0.165      & 0.155    \\ \hline \hline
\multicolumn{1}{c|}{$ODP_{CV}$}  					& 0.583     & 0.550    & 0.536	 & 0.510      & 0.493    \\ \hline
\multicolumn{1}{c|}{$ODP_{WV}$}  					& 0.600     & 0.583    & 0.547	 & 0.508      & 0.482    \\ \hline
\multicolumn{1}{c|}{$ODP_{CV+WV}$}  				& \textbf{0.692}     & \textbf{0.617}    & \textbf{0.583}	 & \textbf{0.556}      & \textbf{0.545}    \\ \hline
\end{tabular}
\end{table}

\vspace*{-0.5cm}

Table \ref{NYT_Performance} shows the evaluation results on the NYT test dataset. Again, the performance of $ODP_{CV+WV}$ outperforms $ODP$, $PV$, $CNN$, $ODP_{CV}$ and $ODP_{WV}$ over 28\%, 119\%, 216\%, 12\%, and 10\% in terms of precision at {\it k} on average, respectively. We also observe that both $ODP_{CV}$ and $ODP_{WV}$ outperform $ODP$. These results clearly demonstrate that both category and word vectors are effective at text classification. Specifically, $ODP_{CV+WV}$, which utilizes both category and word vectors, achieves the best performance in all experiments. We also perform the {\it t}-test for the classification results, and find that $ODP_{CV+WV}$ results are statistically significant with $p$ $\textless$ 0.01.

\vspace*{-0.3cm}

\subsection{Analysis}
We also qualitatively examine the meaning of category vectors to analyze why adding category vectors improves the performance of ODP-based text classification. From Table \ref{wordCheck}, we observe that the category vector expresses the meaning of category quite well. First, from the parent category {\it Home/}{\it Cooking/Baking\_and\_} {\it Confections} and child category {\it Home/}{\it Cooking/Baking\_and\_}{\it Confections/}{\it Breads}, we observe that their category vectors share the core semantically rich words (e.g., Recipe, Baking, Cookies), while they have their own unique semantically rich words (e.g., Dessert, Bread). These observations imply that the category vector actually understands the semantics better than the centroid vector. 

Interestingly, we also observe that the category vector identifies semantically related words that do {\it not} appear in the ODP knowledge base (e.g., Henin, a Belgian former professional tennis player, in the category {\it Sports/Tennis/}{\it Players}). Thus, category vectors combined with the ODP-based classification successfully enable us to improve the performance of text classification.

\vspace*{-0.8cm}

\begin{table}[ht]
\centering
\small
\caption{Nearest words of Category Vector (Explicit + Implicit) and Highly Weighted Words in Centroid Vector (Explicit) of ODP Categories}
\label{wordCheck}
\renewcommand{\arraystretch}{1.0}
\begin{adjustbox}{width=1.0\textwidth}
\begin{tabular}{c|c|c}
\cline{1-3}
\multirow{2}{*}{Category}  & Nearest Words of Category Vector & Highly Weighted Words in Centroid Vector\\
                                   		& (Explicit + Implicit) & (Explicit)\\\hline
\hline
$Home/Cooking/Baking\_$  & Recipe, Baking, Cookies, Cake & Recipe, Baking, Cookies, Cake\\
$and\_Confections$ & Dessert, Cupcake, Bake, ... & Bake, Pastries, Bread, Mix, ...\\
\hline
$Home/Cooking/Baking\_$  & Bread, Recipe, Baking, Flour, & Bread, Recipe, Sourdough,\\
$and\_Confections/Breads$ & Biscuit, Cookies, Pancake, ... & Baking, Yeast, Quick, ...\\
\hline
\multirow{2}{*}{$Sports/Tennis/Players$}  & Tennis, Wimbledon, Nadal, & Tennis, Wimbledon, Winners,\\
                                   		& Henin, Federer, Sharapova, ... & Players, Detailed, Seed, ...\\\hline
\end{tabular}
\end{adjustbox}
\end{table}

\vspace*{-0.9cm}

\section{Related Work}
\vspace*{-0.2cm}
For the large-scale text classification, many approaches have been developed to handle data sparsity on a knowledge base. Data sparsity on a hierarchical taxonomy was firstly addressed in \cite{McCallum:Shrinkage}. This work applied a statistical technique to estimate the parameters of data-sparse child categories with their data-rich ancestor categories. In \cite{Lee:MC-TWEB,Ha:MCAD}, they proposed the merge-centroid (MC) classification that utilizes enriched training data for each category based on webpages classified into their ancestor and/or descendants in the ODP. In another line of work \cite{Shin:Wikipedia}, they enriched semantic information in the ODP by incorporating another knowledge base, Wikipedia.

A simple convolutional neural network approach \cite{Kim:CNN} has been proven to be an effective text classifier. Still, it exhibits limitations in the large-scale text classification, which is verified in our analysis. A few work \cite{Nam:LMT,Gakuto:INN} has recently studied large-scale multi-label text classification using deep neural networks. However, they do not utilize the explicit representation model built from knowledge base. To the best of our knowledge, our current work is one of only a few works that utilizes both the explicit and implicit knowledge representation, which enables us to perform the large-scale text classification quite well.  

\vspace*{-0.2cm}

\section{Conclusion}
\vspace*{-0.2cm}
In this paper, we have proposed novel joint models of the explicit and implicit representation techniques to handle the large-scale text classification. Specifically, we have incorporated the well-known word2vec model into the ODP-based classification framework. Our approach involves two tasks. First, we generate category vectors, which represent the semantics of ODP categories. Second, we develop a new semantic similarity measure that utilizes both category and word vectors. We have verified the large-scale classification performance of the proposed methodology using real-world datasets. The performance evaluation results confirm that our scheme significantly outperforms baseline methods. We plan to apply the proposed methodology to different applications, including contextual and mobile advertising.

\vspace*{-0.3cm}

\section*{Acknowledgment}
\vspace*{-0.2cm}
This research was supported by Basic Science Research Program through the National Research Foundation of Korea (NRF) funded by the Ministry of Science, ICT (numbers 2015R1A2A1A10052665 and 2017M3C4A7077601).

\vspace*{-0.2cm}

\bibliographystyle{splncs} 
\bibliography{typeinst}

\end{document}